\pdfoutput=1

\documentclass[11pt]{article}

\usepackage{EMNLP2022}

\usepackage{times}
\usepackage{latexsym}
\usepackage{algorithmic}
\usepackage[]{algorithm2e}
\usepackage{placeins}
\usepackage{xcolor}
\usepackage{comment}
\usepackage{amsmath}
\usepackage{multirow}
\usepackage{tabularx,booktabs}
\usepackage{todonotes}

\renewcommand{\cite}[1]{\citep{#1}}

\usepackage{amsmath}

\usepackage[T1]{fontenc}

\usepackage[utf8]{inputenc}

\usepackage{microtype}

\usepackage{inconsolata}

\newcommand{\best}[1]{\textbf{#1}}
\newcommand{\sbest}[1]{\underline{#1}}

\newcommand{\retwog}{$\text{Re}^2\text{G}$}
\newcommand{\kgi}[1]{$\text{KGI}_{#1}$}

\title{KGI: An Integrated Framework for Knowledge Intensive Language Tasks}

\author{
    Md Faisal Mahbub Chowdhury$^*$, 
    Michael Glass$^*$, 
    Gaetano Rossiello,\\
    \bf Alfio Gliozzo \and
     Nandana Mihindukulasooriya
    \\
IBM Research AI, Yorktown Heights, NY, USA
}

\begin{document}
\maketitle

\def\thefootnote{*}\footnotetext{These authors contributed equally to this work.}\def\thefootnote{\arabic{footnote}}

\begin{abstract}
In this paper, we present a system to showcase the capabilities of the latest state-of-the-art retrieval augmented generation models trained on knowledge-intensive language tasks, such as slot filling, open domain question answering, dialogue, and fact-checking. 
Moreover, given a user query, we show how the output from these different models can be combined to cross-examine the outputs of each other. 
Particularly, we show how accuracy in dialogue can be improved using the question answering model. We are also releasing all  models used in the demo as a contribution of this paper. A short video demonstrating the system is available at \url{https://ibm.box.com/v/emnlp2022-demo}. 
\end{abstract}

\section{Introduction}



Recently, we proposed \retwog~\cite{glass-etal-2022-re2g}, the core of our \kgi{} (Knowledge Graph Induction) system.
\retwog~combines both neural initial retrieval and reranking into a BART-based sequence-to-sequence generation.
We show that the end-to-end reranking component also permits merging retrieval results from sources with incomparable scores, enabling an ensemble of BM25 and neural initial retrieval.
Moreover, to train our system end-to-end, we introduce a novel variation of knowledge distillation to train the initial retrieval, reranker, and generation using only ground truth on the target sequence output. 
We find large gains in four diverse tasks: zero-shot slot filling, question answering, fact-checking, and dialog, with relative gains of 9\% to 34\% over the previous state-of-the-art on the KILT leaderboard~\cite{kilt}\footnote{https://eval.ai/web/challenges/challenge-page/689/overview}.

 
 
In this work, we describe the complete KGI system, which is an enhancement of our previous work. We demonstrate how users can asynchronously interact with the system in real-time, not only for completing triples (aka slot filling), but also for dialogue, fact-checking, and open-domain question answering. We empirically show that our system is the state of the art for these tasks on the KILT leaderboard. In addition, we show how dialog accuracy can be improved by exploiting the question answering model, a novel approach demonstrated in this paper.
 
There are several different intended usages of our system. For example, KGI allows users to interact with different levels of verbosity. Also, it enables users to cross-examine results through different KILT tasks that are part of the same GUI.

We are releasing our best KGI core models (i.e. \retwog) that we used in this paper at \url{https://huggingface.co/ibm} .

\section{System Architecture}


\begin{figure}[t]
  \centering
  \includegraphics[width=.9\linewidth]{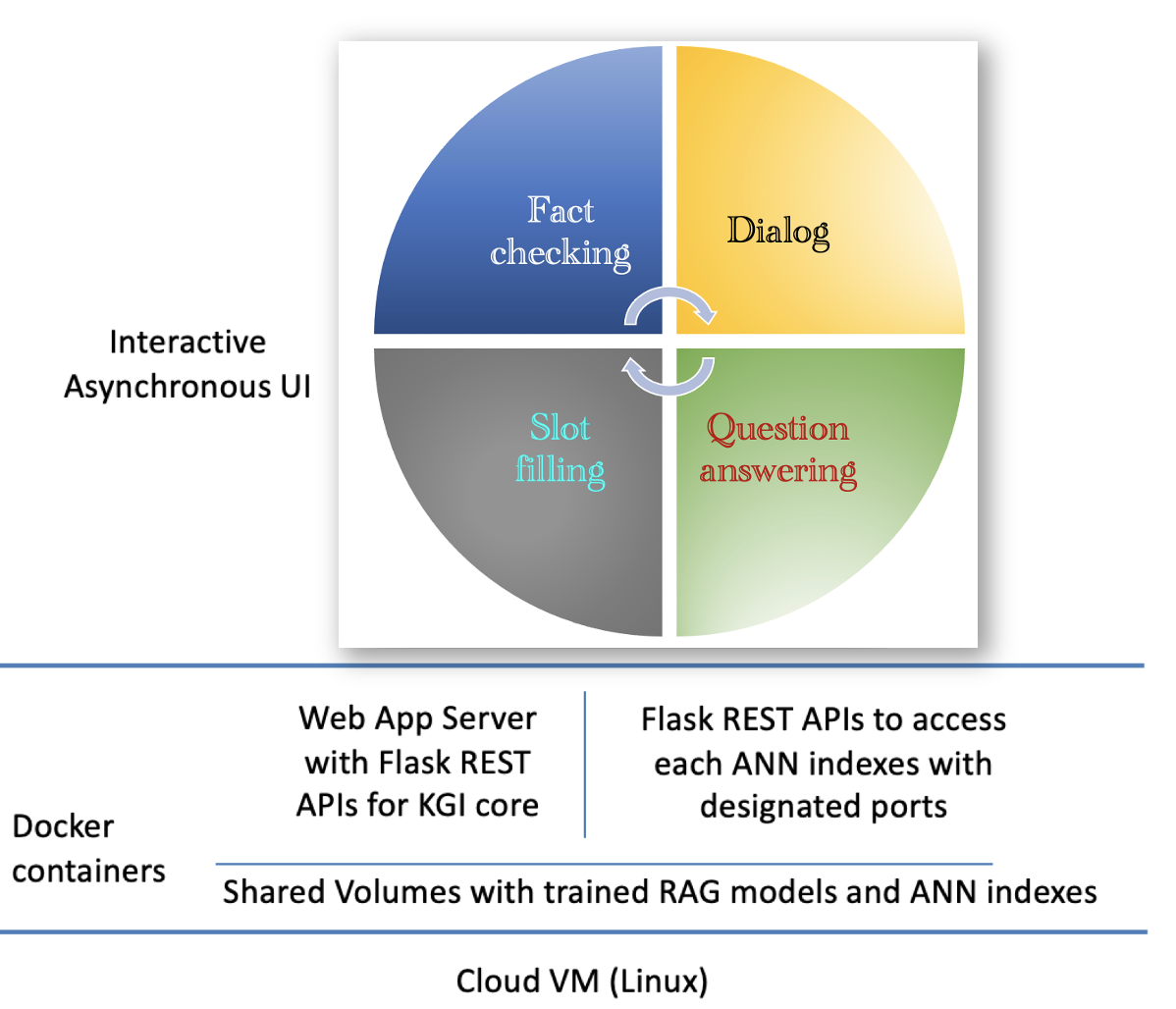}
  \caption{KGI: System Architecture}
  \label{fig.sys}
\end{figure}

\begin{figure*}[t]
  \centering
  \includegraphics[width=.9\linewidth]{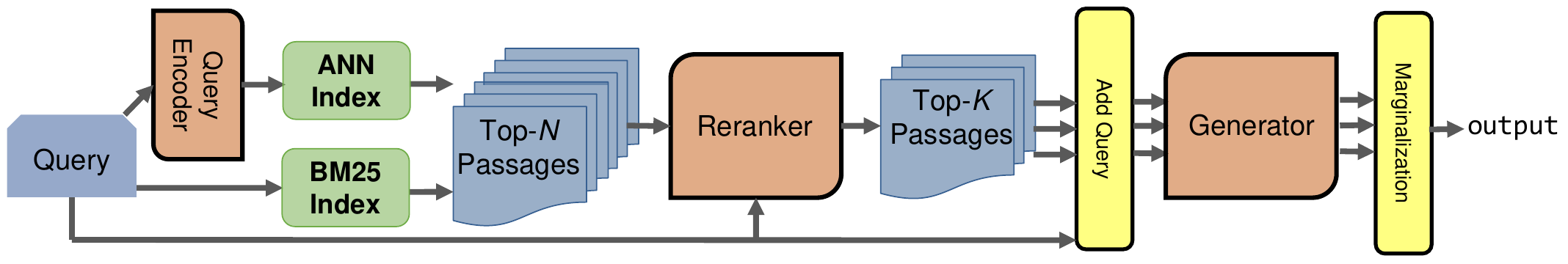}
  \caption{Re2G (Retrieve, Rerank, Generate) model~\cite{glass-etal-2022-re2g}}
  \label{fig.re2g}
\end{figure*}

The KGI system (Figure~\ref{fig.sys}) is a web-based application that enables users to asynchronously interact with the system in real-time and allows users to obtain results from four different task-specific models simultaneously in different tabs in the GUI. These models are trained using \retwog~model as shown in Figure~\ref{fig.re2g}. There is a corresponding ANN (Approximate Nearest Neighbors) index, in this case HNSW (Hierarchical Navigable Small World) \cite{hnsw} using the open source FAISS library \cite{faiss}\footnote{\url{https://github.com/facebookresearch/faiss}}. These indexes contain the passage vectors for the source corpus of the corresponding tasks. More details about the KGI core models, particularly how a model can be trained and a corresponding index can be created, can be found in~\citet{glass-etal-2021-robust, glass-etal-2022-re2g}.

The KGI core model, such as \retwog, takes a textual query as input and returns a set of generated texts as results together with a set of passages as supporting evidences/references for each of the tasks such as slot filling or fact-checking. Refer to Table \ref{tbl.application} for examples.


\subsection{Dialog supported by QA}
\label{subsec.dialog_hybrid}
There are two settings for dialog from the GUI. \textit{``conventional-dialog''} is solely based on the KGI dialog model, while in the \textit{``hybrid''} settings the system also interacts with the KGI QA model depending on the comments entered by the user. The system uses a simple Convolutional Neural Networks (CNN) based text classification model to detect whether the latest comment entered by the user is a question. If a comment is identified as a question, and if it contains at least one noun phrase without a pronoun  or adverb, the system creates a query by appending all such noun phrases from the previous user utterances in the current dialog history (with full-stop as separators) with the question and pass it to the QA model. If none of the tokens in the best-ranked answer provided by the QA model is part of the same dialog history, the system picks the QA answer (and corresponding evidences) as the response for the dialog.

\section{Application to Diverse NLP Tasks}
\label{application_to_other_tasks}

\subsection{The tasks}

As mentioned earlier, we demonstrate the robustness of our system on four NLP tasks that are part of the KILT leaderboard. Among them, fact-checking requires deep knowledge about the claim and reasoning over multiple documents. In slot filling, the goal is to collect information on certain relations of entities. For the open domain QA, the goal is producing the correct answer for a question after reasoning over an entire knowledge source (in this case, Wikipedia), without a predefined location for the answer. Finally, for dialog the goal of the system is to engage in chitchat, relying on topical and factual knowledge, on a wide array of (non-specified) topics with a user. There is another task, entity linking, in KILT in which we did not participate.

The KILT benchmark consists of eleven datasets spanning five distinct tasks. All these task-specific datasets in this benchmark are grounded in the same snapshot of Wikipedia. We refer the readers to \citet{kilt} for details about the datasets. Table \ref{tbl.application} shows the input and output types for the four different tasks considered.

\subsection{Application of KGI}

The KGI system, although originally designed for zero-shot slot filling, is based on a very general approach: conditional generation with retrieval.  An input text is used to retrieve passages from a corpus of knowledge, then a generation component conditions both the input text and the returned passages to produce an output text. 

The KILT benchmark was introduced to evaluate the capabilities of pre-trained language models to address NLP tasks that require access to external knowledge. As mentioned by the organizers, developing general models for such knowledge-intensive tasks is difficult as each task might require computationally expensive indexing of custom knowledge sources, in addition to dedicated infrastructure. So, it is a perfect playground to verify the generalizability and robustness of KGI.

Training models for each of the above tasks are carried out in two phases: DPR training and generation training. The training procedure and hyperparameters are exactly the same as described in our earlier works \cite{glass-etal-2021-robust,glass-etal-2022-re2g}\footnote{\url{https://github.com/IBM/kgi-slot-filling}}\footnote{\url{https://github.com/IBM/kgi-slot-filling/tree/re2g}}. 

The slot filling dataset, T-REx \cite{trex}, provides as input a head entity and relation, and expects as output the entity or term that fills the slot, also called the tail entity.   
The T-REx dataset contains 2.3M instances. 
We use only 370k training instances by down-sampling the relations that occur more than 5000 times. This reduces the training time required while keeping state-of-the-art performance. The development and test sets each have 5k instances.

The question answering datasets are ``open'' versions of Natural Questions~\cite{naturalquestions} and TriviaQA~\cite{triviaqa}. Unlike the original versions, the relevant Wikipedia page must be found by a retrieval step. The training sets for Natural Questions and TriviaQA contain 87k and 62k questions, with another 3k and 5k for the development and 1.4k and 6.5k for test.

The fact-checking dataset in KILT is FEVER (Fact Extraction and VERification). It is a combination of the two FEVER versions \cite{fever,fever2}  omitting the \textsc{NotEnoughInfo} class. There are approximately 10k instances in the development and test sets, and 100k for training. FEVER is a classification task, but we cast it as a generation task by training the model to generate either the token ``SUPPORTS'' or ``REFUTES''.

Wizard of Wikipedia \cite{wizardofwikipedia} is the dialog dataset. The input is a short dialog history ending with the information seeker's turn. The expected output is a fact presented conversationally or just an utterance or question mentioning content from a relevant Wikipedia page.  It is the smallest dataset with approximately 3k instances in development and test and 64k in train.

\begin{table*}
\begin{center}
\small
\renewcommand{\arraystretch}{1.5} 
\begin{tabular}{p{1.4cm}p{1.5cm}p{1.3cm}p{5cm}p{1.5cm}p{3cm}}
\textbf{Task} & \textbf{Dataset} & \textbf{Input} & \textbf{Example} & \textbf{Output} & \textbf{Example} \\
\hline
Slot filling & T-REx & Head [SEP] Relation & \texttt{Elizabeth Cromwell [SEP] spouse} & Tail Entity & \texttt{Oliver Cromwell} \\ \hline
Fact \ \ checking & FEVER & Claim \ \ sentence & \texttt{Slovenia uses the euro.} & Truth Classification & \texttt{SUPPORTS} \\ \hline
Dialog & Wizard of Wikipedia & Dialog history & \texttt{
... Those sound wonderful. Can you tell me any more information? * Iceland is sparsely populated and in fact has the smallest population in Europe. * What other countries are around it?} & Next dialog turn & \texttt{Denmark, Iceland, Finland, Norway and Sweden are all Nordic countries.} \\ \hline

Question \ \ Answering & TriviaQA, \ \ Natural \ \ Questions & Question & When did bram stoker’s dracula come out? & Answer & \texttt{1987} \\ \hline

\end{tabular}
\end{center}
\caption{Application of conditional generation with retrieval to KILT tasks}
\label{tbl.application}
\end{table*}

\subsection{Results}

Table \ref{tbl.experimentalResults} shows the results of our system on  KILT datasets for different tasks. At the time of the submission in 2021, our earlier version of KGI core models (namely, \kgi{0} and \kgi{1}) achieved the best results in the KILT leaderboard. Our new KGI core models, \retwog, achieves significantly better results. In fact, it considerably advanced the state-of-the-art across five KILT datasets, and still holds the top position in four of the five. Particularly, our system architecture permits us to ensemble DPR and BM25, which is enabled by our incorporation of a reranker, further improving accuracy. Our online knowledge distillation is able to improve the performance of DPR in four of the five datasets, despite the loss in end-to-end training not depending on the DPR scores.

\begin{table*}
\begin{center}
\small
\renewcommand{\arraystretch}{1.2} 
\begin{tabular}{r|cc|ccccc}
& \multicolumn{6}{c}{\textbf{T-REx}}\small(Slot Filling) \\ \hline
 & \textbf{R-Prec}  & \textbf{Recall@5} & \textbf{Accuracy} & \textbf{F1}  & \textbf{KILT-AC} & \textbf{KILT-F1}\\
\hline 
\retwog~\cite{glass-etal-2022-re2g}      & \best{80.70} & \best{89.00} & \best{87.68} & \best{89.93} & \best{75.84} & \best{77.05} \\
\kgi{1}~\cite{glass-etal-2021-robust}    & 74.36 & 83.14 & \sbest{84.36} & \sbest{87.24} & \sbest{69.14} & \sbest{70.58} \\
KILT-WEB 2~\cite{kilt_web2}   & \sbest{75.64} &	\sbest{87.57} & 81.34 & 84.46 & 64.64 & 66.64 \\
SEAL~\cite{seal_kilt}         & 67.80 & 81.52 & 83.72 & 86.53 & 60.08 & 61.72 \\
\kgi{0}~\cite{glass-etal-2021-robust}      & 59.70 & 70.38 & 77.90 & 81.31 & 55.54 & 56.79 \\
\hline 

& \multicolumn{6}{c}{\textbf{Natural Questions}}\small(Question Answering) \\ \hline
 & \textbf{R-Prec}  & \textbf{Recall@5} & \textbf{Accuracy} & \textbf{F1}  & \textbf{KILT-AC} & \textbf{KILT-F1}\\
\hline 
\retwog~\cite{glass-etal-2022-re2g}           & \best{70.78} & \best{76.63} & \sbest{51.73} & \sbest{60.97} & \best{43.56} & \best{49.80} \\
SEAL~\cite{seal_kilt}    & 63.16 & 68.19 & \best{53.74} & \best{62.24} & \sbest{38.78} & \sbest{44.40} \\
\kgi{0}~\cite{glass-etal-2021-robust} & \sbest{63.71} & 70.17 & 45.22 & 53.38 & 36.36 & 41.83 \\
KILT-WEB 2~\cite{kilt_web2}   & 59.83 & \sbest{71.17} & 51.59 & 60.83 & 35.32 & 40.73 \\
RAG~\cite{kilt}          & 59.49 & 67.06 & 44.39 & 52.35 & 32.69 & 37.91 \\
\hline 

& \multicolumn{6}{c}{\textbf{TriviaQA}}\small(Question Answering) \\ \hline
 & \textbf{R-Prec}  & \textbf{Recall@5} & \textbf{Accuracy} & \textbf{F1}  & \textbf{KILT-AC} & \textbf{KILT-F1}\\
\hline 
\retwog~\cite{glass-etal-2022-re2g}          & \best{72.68} & \sbest{74.23} & \best{76.27} & \best{81.40} & \best{57.91} & \best{61.78} \\
SEAL~\cite{seal_kilt}    & \sbest{68.36} & \best{76.36} & 70.86 & 77.29 & \sbest{50.56} & \sbest{54.99} \\
KILT-WEB 2~\cite{kilt_web2}  & 58.85 & 71.55 & \sbest{72.73} & \sbest{79.54} & 45.55 & 49.57 \\
\kgi{0}~\cite{glass-etal-2021-robust} & 60.49 & 63.54 & 60.99 & 66.55 & 42.85 & 46.08 \\
MultiDPR~\cite{multidpr} & 61.49 & 68.33 & 59.60 & 66.53 & 42.36 & 46.19 \\
\hline 

& \multicolumn{6}{c}{\textbf{FEVER}} \small(Fact Checking) \\ \hline
 & \textbf{R-Prec}  & \textbf{Recall@5} & \textbf{Accuracy} &   & \textbf{KILT-AC} & \\
\hline
\retwog~\cite{glass-etal-2022-re2g}     & \best{88.92} & \best{92.52} & \best{89.55} &  & \best{78.53} & \\
SEAL~\cite{seal_kilt}    & \sbest{81.45} & \sbest{89.56} & \sbest{89.54} & & \sbest{71.28} & \\
KILT-WEB 2~\cite{kilt_web2}  & 74.77 & 87.89 & 88.99 & & 65.68 & \\
\kgi{0}~\cite{glass-etal-2021-robust}     & 75.60 & 84.95 & 85.58 &  & 64.41 & \\
MultiDPR~\cite{multidpr} & 74.48 & 87.52 & 86.32 & & 63.94 & \\

\hline 
& \multicolumn{6}{c}{\textbf{Wizard of Wikipedia}} \small(Dialog) \\ \hline
 & \textbf{R-Prec}  & \textbf{Recall@5} & \textbf{Rouge-L} & \textbf{F1}  & \textbf{KILT-RL} & \textbf{KILT-F1}\\
\hline 
Hindsight~\cite{hindsight_arxiv} & 56.08 & 74.27 & \best{17.06} & \best{19.19} & \best{11.92} & \best{13.39} \\
\retwog~\cite{glass-etal-2022-re2g}      & \best{60.10} & \best{79.98} & \sbest{16.76} & \sbest{18.90} & \sbest{11.39} & \sbest{12.98} \\
SEAL~\cite{seal_kilt}    & 57.55 & \sbest{78.96} & 16.65 & 18.34 & 10.45 & 11.63 \\
\kgi{0}~\cite{glass-etal-2021-robust}      & 55.37 & 78.45 & 16.36 & 18.57 & 10.36 & 11.79 \\
RAG~\cite{kilt}          & \sbest{57.75} & 74.61 & 11.57 & 13.11 & 7.59 & 8.75 \\
KILT-WEB 2~\cite{kilt_web2}  & 41.54 & 68.25 & 13.94 & 15.66 & 6.55 & 7.57 \\
\end{tabular}
\end{center}
\caption{KILT leaderboard top systems. \retwog~\cite{glass-etal-2022-re2g}, \kgi{0}~\cite{glass-etal-2021-robust} and \kgi{1}~\cite{glass-etal-2021-robust} are different KGI core models from our recent work.}
\label{tbl.experimentalResults}
\end{table*}

\section{Examples and Analysis}
\label{sec:analysis}

\subsection{Complementing information from different applications}
\begin{figure*}[t]
  \centering
  \includegraphics[width=1\linewidth]{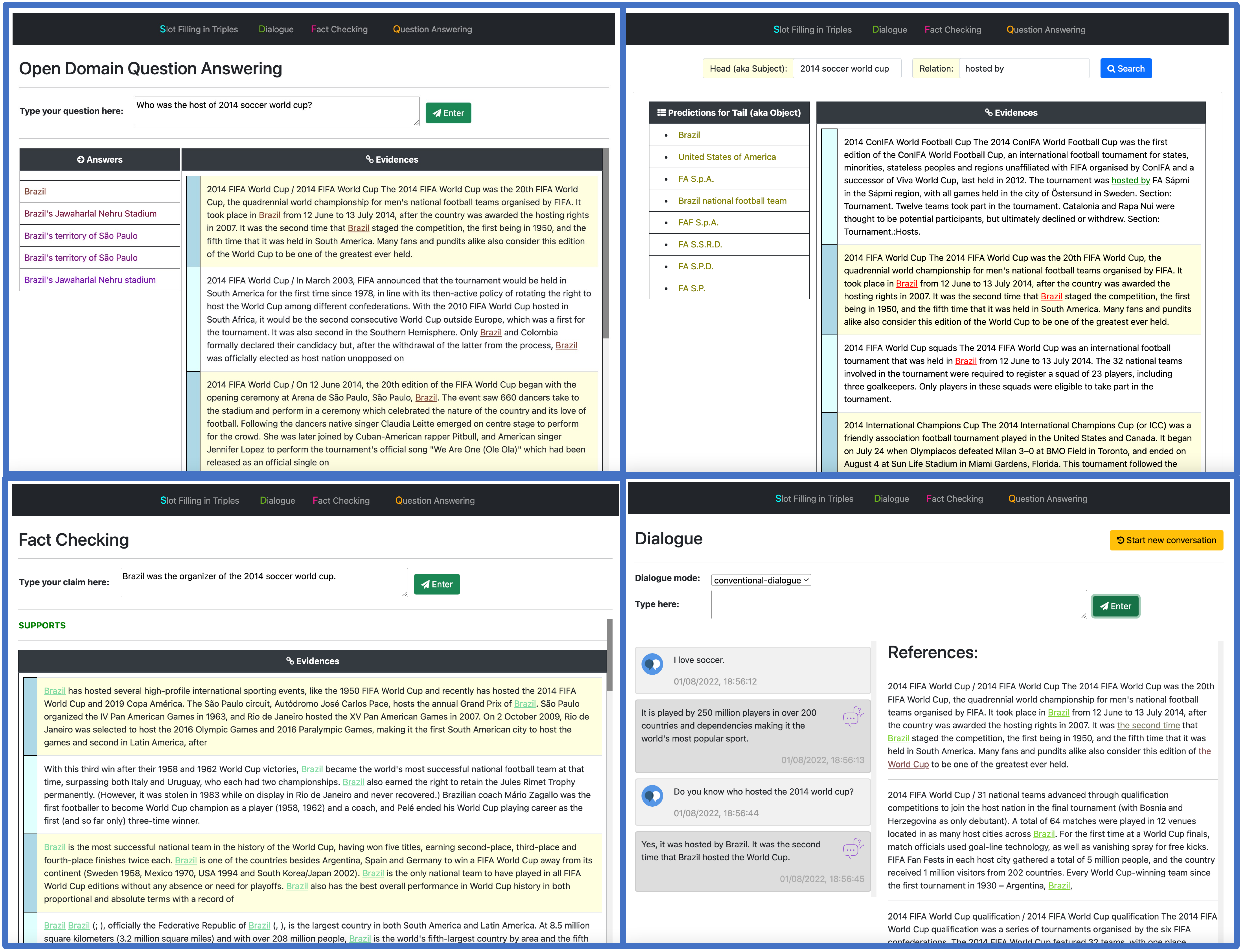}
  \caption{Response from different task models of our system for the same common query}
  \label{fig.example}
\end{figure*}

As mentioned earlier, one of our goals is to allow a user to interact with different levels of verbosity and then cross-examine the results to check whether response from one application (e.g. fact checking) supports response from another application (e.g. dialog). This can be checked not only by looking at the responses but also through the accompanying evidences. Figure \ref{fig.example} shows such an example where the user intends to know the host of the 2014 Soccer World Cup. This has been formulated in different ways according to the application. All four KGI models for the corresponding applications provided the correct answer (Brazil). Note, all the models use the Wikipedia corpus (as provided by the KILT organizers) as the knowledge source, yet the corresponding supporting evidences are not always the same.

In real world scenarios, the ability to cross-check information and compare complementing evidence is important for  decision making, specially for subject matter experts.

\subsection{Dialog by exploiting results of open domain QA}

\begin{table*}[!h]
\begin{center}
\small
\renewcommand{\arraystretch}{1.5}
\begin{tabular}{ll}
\textbf{Conversation 1:} & \\
\texttt{Human:} & I think a lot of young people are addicted to social media platforms.\\
\texttt{KGI (both dialog):} & I think so too. Social media can be used to promote a product or service, but it can  \\
& also be used as a way to get information and ideas.\\
\texttt{Human:} & I sometimes check Facebook and post photos there but I don't use it very often.\\
\texttt{KGI (both dialog):} & I don't use it as much myself, but I do see it as a way to get information and ideas.\\
\texttt{Human:} & Do you know when was Facebook first launched?\\ \hline
\texttt{KGI (hybrid dialog):} & February 4, 2004 .\\ \hline
\texttt{KGI (conventional dialog):} & I'm not sure, but I do know that it was founded by Mark Zuckerberg in the 1980s.\\ \hline \hline

\textbf{Conversation 2:} & \\
\texttt{Human:} & I use my mobile phone all the time, it is really useful.\\
\texttt{KGI (both dialog):} & I love my mobile phone, it is a portable telephone that can make and receive calls over a \\
& radio frequency link while the user is moving within a telephone service area. \\
\texttt{Human:} & Yes, I can't imagine how we managed to work together when we didn't have mobile phones.\\
\texttt{KGI (both dialog):} & Modern mobile telephone services use a cellular network architecture, and, therefore, \\
& mobile telephones are called "cellular telephones" or "cell phones" in North America.\\
\texttt{Human:} & Interesting. Who is the inventor of cell phones?\\ \hline
\texttt{KGI (hybrid dialog):} & William Rae Young, Jr.\\ \hline
\texttt{KGI (conventional dialog):} & I am not sure, but I do know that the first handheld mobile phone was envisioned \\
& in the early stages of radio engineering.
\\ \hline  \hline

\textbf{Conversation 3:} & \\
\texttt{Human:} & I ate two doughnuts for breakfast today. \\
\texttt{KGI (both dialog):} & I love doughnuts. They are a type of fried dough confectionery. \\
\texttt{Human:} & Yes, they are so delicious. I love them too. \\
\texttt{KGI (both dialog):} & The doughnut is popular in many countries and prepared in various  forms as a sweet  \\
& snack that can be homemade or purchased in bakeries, supermarkets, food stalls,  \\
& and franchised specialty vendors. \\
\texttt{Human:} & What are the main ingredients used to make doughnuts?\\ \hline
\texttt{KGI (both dialog):} & They are usually deep fried from a flour dough, and typically either ring-shaped or a  \\ & number of shapes without a hole.\\ \hline \hline

\textbf{Conversation 4:} & \\
\texttt{Human:} & Harry Potter books have been translated into more than 80 languages. \\
\texttt{KGI (both dialog):} & The first book, Harry Potter and the Philosopher's Stone, has been translated into over \\
& 75 languages. \\
\texttt{Human:} & I see. I think all the books in the series are really popular all over the world. \\
\texttt{KGI (both dialog):} & Yes, they are very popular. The first book, Harry Potter and the Philosopher's Stone, \\
& was published in 1997. \\
\texttt{Human:} & What's the title of the last book?\\ \hline
\texttt{KGI (both dialog):} & It's called the Deathly Hallows. It was released on 21 July 2007.\\ \hline

\end{tabular}
\end{center}
\caption{\centering Examples of conversations done by the human evaluator to compare different KGI dialog settings. The responses provided by the hybrid dialog in the first 2 conversations are different and better.}
\label{tbl.different_dialogs}
\end{table*}

Our view is that the most natural choice to automatically combine results from different task specific models and improve results of a particular task is dialog. So, we created a hybrid settings for the dialog application as described in Section \ref{subsec.dialog_hybrid}. 


We asked an experienced AI researcher (whose background is not NLP and who was not involved in building this system) to be a user of our system and compare the hybrid and standalone/conventional dialog settings. We gave the user the following instructions --

\begin{itemize}
    \item The user will perform 20 independent conversations.
    \item The user can chat about anything he likes.
    \item The user should not make the topic of the conversation explicit to the system. We wanted the system to understand it from the conversation.
    \item The user should limit his interactions to 3 turns, where the first and second utterances by the user will be followed by a question. This ensures that in both of the settings, the system has the same context for the conversation. Note, the QA model is only exploited by the system during a dialog when a question is asked.
    \item The question in each of the conversation should be a factoid question.
    \item At the end of each conversation, the user will mark which of the dialog settings provided a better factually correct response.
\end{itemize}

According to the user, in \textbf{10 out of 20} conversations the hybrid settings provided better factual results. In the rest of the 10 conversations, the responses were the same, i.e. the system opted for the output generated by the dialog model. To put it differently, in this limited human evaluation, in the hybrid settings whenever the system choose the QA model generated response, it was always correct. Table \ref{tbl.different_dialogs} shows few examples of conversations conducted by the human user.

\section{Conclusions}

In this work, we present our KGI system and show how a user can asynchronously interact with it in real time simultaneously for four NLP tasks. This allows users to interact with same system with different levels of verbosity. The KGI core models (\retwog) is still the state of the art (in some cases, by wide margin) for 4 out these 5 tasks in the KILT leaderboard. We are releasing those models to the community as part of this paper.

In addition, we show how dialog accuracy can be improved by exploiting open domain QA where both models are grounded in the same snapshot of knowledge source. In future work, we would like to exploit results from fact checking and slot filling to further improve accuracy of the response in dialog. We have also directed our efforts towards improving the retrieval of relevant knowledge which would enable improvement in end-to-end performance by supplying better passages to the generation component.

\FloatBarrier

\bibliography{anthology,custom}
\bibliographystyle{acl_natbib}

\end{document}